\documentclass[letterpaper, 10 pt, conference]{IEEEtran}
\IEEEoverridecommandlockouts

\let\labelindent\relax
\usepackage{enumitem}
\usepackage{balance}
\usepackage[font={small}]{caption}
\usepackage{subcaption}
\usepackage{array}
\usepackage{textcomp}
\usepackage{mathtools, nccmath}
\usepackage{graphicx}
\usepackage{amsfonts}
\usepackage{amsmath}
\usepackage{amssymb}
\usepackage{algorithm}
\usepackage{algpseudocode}
\usepackage{hyperref}
\usepackage{tikz}
\usepackage{arydshln}
\usepackage{multirow}
\usepackage{bm}
\usepackage{epstopdf}
\usepackage{cite}
\usepackage{siunitx}
\usepackage{bbm}
\usepackage{diagbox}
\usepackage{soul}
\usepackage{makecell}
 
\DeclareMathOperator*{\argmin}{argmin} 

\newtheorem{remark}{Remark}

\begin{document}

\title{\LARGE \bf IteraOptiRacing: A Unified Planning-Control Framework for Real-time Autonomous Racing for Iterative Optimal Performance}
 
\author{Yifan Zeng$^{*1}$, Yihan Li$^{*2}$, Suiyi He$^3$, Koushil Sreenath$^4$, Jun Zeng$^4$
\thanks{
$^*$Authors have contributed equally.
}
\thanks{$^1$Author is with Shanghai Jiao Tong University, Shanghai, China. {\tt\small blakezyf1107@sjtu.edu.cn}}
\thanks{$^2$Author is with University of Pennsylvania, Philadelphia, USA. {\tt\small yianlee@seas.upenn.edu}}
\thanks{$^3$Author is with University of Minnesota-Twin Cities, MN 55455, USA. {\tt\small he000231@umn.edu}}
\thanks{$^4$Authors are with University of California, Berkeley. {\tt\small \{koushils, zengjunsjtu\}@berkeley.edu}}
}

\maketitle
\begin{abstract}
This paper presents a unified planning-control strategy for competing with other racing cars called IteraOptiRacing in autonomous racing environments.
This unified strategy is proposed based on Iterative Linear Quadratic Regulator for Iterative Tasks (i2LQR), which can improve lap time performance in the presence of surrounding racing obstacles. 
By iteratively using the ego car's historical data, both obstacle avoidance for multiple moving cars and time cost optimization are considered in this unified strategy, resulting in collision-free and time-optimal generated trajectories.
The algorithm's constant low computation burden and suitability for parallel computing enable real-time operation in competitive racing scenarios.
To validate its performance, simulations in a high-fidelity simulator are conducted with multiple randomly generated dynamic agents on the track.
Results show that the proposed strategy outperforms existing methods across all randomly generated autonomous racing scenarios, enabling enhanced maneuvering for the ego racing car.
\end{abstract}

\IEEEpeerreviewmaketitle
\section{Introduction}
\label{sec:introduction}
\subsection{Motivation}
\begin{figure}[t]
    \setlength{\abovecaptionskip}{0.5cm}
    \setlength{\belowcaptionskip}{-0.5cm}
    \centering
    \includegraphics[width=0.9\linewidth]{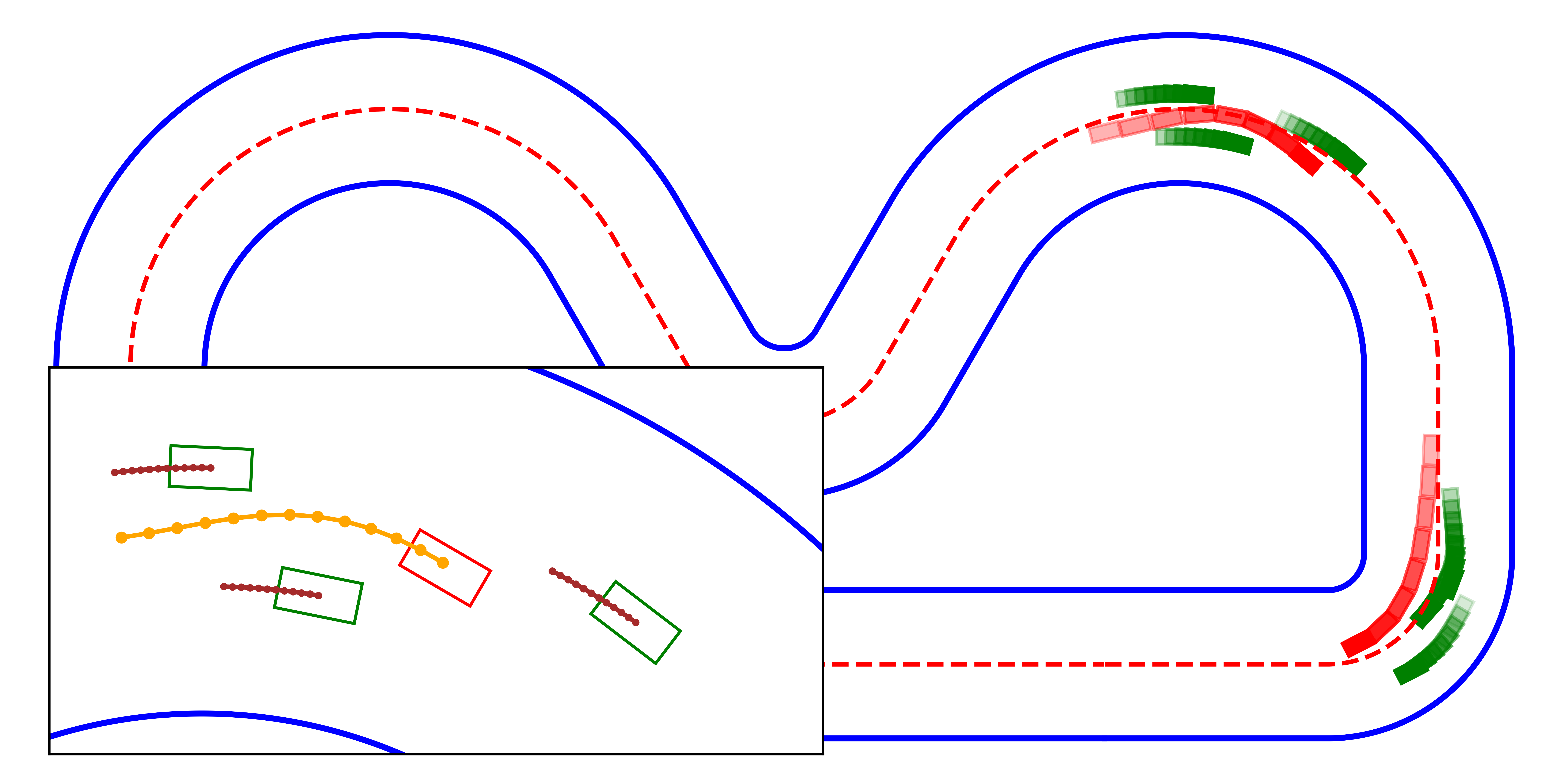}
    \caption{Snapshots from two simulation cases of the overtaking behavior in a m-shape track.
    The red rectangle represents the ego vehicle, and the green rectangles represent the moving vehicles that serve as obstacles.   
    Note that the proposed racing algorithm enables the ego vehicle to smoothly overtake other racing cars.
    The solid blue line and dashed brown line indicate the track's boundary and center line, respectively.
    The orange line shown in the inset is the optimized open-loop predicted trajectory, and the brown line represents the prediction trajectories of the moving vehicles acting as obstacles.}
    \label{fig:introduction}
\end{figure}

Recently, there has been a growing interest in autonomous racing \cite{wischnewski2022indy, betz2205tum,song2023reaching, sun2023benchmark}, which is a challenging subtopic in the field of autonomous driving research.
In such racing competitions, the ego vehicle is expected to complete the required number of laps on a designated track in the shortest time possible. 
To achieve this goal, the autonomous racing algorithm must address two critical challenges: maximizing driving speed while simultaneously competing with other cars on the same track.
Traditionally, most existing work in this area tackles these two problems separately.
However, to secure victory in a race, an algorithm must deliver time-optimal behavior in the presence of other competing dynamic vehicles. 
In response to this need, we propose a racing algorithm that enables the ego vehicle to maintain high-speed performance even in the presence of surrounding competing vehicles by considering global optimality, as shown in Fig.~\ref{fig:introduction}.

\subsection{Related Work}

In contrast to the planning and control algorithms for autonomous vehicles on public roads, autonomous racing algorithms must push the vehicle to its dynamic limit \cite{wischnewski2019model}.
When competing on a track with other vehicles, the racing car must overtake rivals as quickly as possible while adhering to the constraints of track boundaries and ensuring safety.  
These capabilities are crucial for achieving the fastest lap times and ultimately winning the race \cite{betz2022autonomous, wischnewski2022indy}.
However, these objectives are inherently challenging.
Calculating the optimal trajectory for a complex race track can be time-consuming, especially in the presence of other moving vehicles.
Additionally, as the vehicle approaches its physical limits, a more accurate dynamics model is required by the planning and control algorithm. 
Furthermore, given the rapidly changing racing environment, autonomous racing algorithms must update at a high frequency while maintaining a long prediction horizon to ensure optimal performance.
Any significant time delays caused by the complex nonlinear optimization can severely compromise the strategy's effectiveness~\cite{kalaria2023delay}.
Previous studies provide valuable insights into this field. 
Below, we review related work across diverse fields using different methodologies.

Model-based planning and control algorithms are widely used in autonomous racing competitions.
A common strategy is to divide the task into two subtasks: minimizing lap times and overtaking leading vehicles, with autonomous racing algorithms tackling these issues separately. 
To minimize lap times, some projects employ low-level controllers to track the time-optimal trajectory of the race track.
Optimization-based approaches are widely used to generate these time-optimal race lines \cite{heilmeier2019minimum,bonab2019optimization,caporale2019towards,alcala2020autonomous,herrmann2020real,srinivasan2021holistic, numerow2024inherently}.
Since autonomous racing cars always drive at their physical limit, improving the low-level controller's tracking performance in these scenarios is crucial \cite{laurense2017path, hewing2018cautious, zhang2018tire}.
Meanwhile, reference-free model-based algorithms, such as Model Predictive Contouring Control (MPCC) \cite{kabzan2019learning} with dynamics learning, and data-driven based Iterative Learning Control (ILC) \cite{rosolia2019learning, kapania2020learning}, also demonstrate strong potential for achieving exceptional performance in autonomous racing.

However, above approaches are limited to scenarios without obstacles or with only static obstacles, making them more suitable for pole position competitions during qualifying sessions.
On race day, autonomous racing cars face a more complex environment, competing with multiple racing cars. 
This requires quick and safe overtaking maneuvers in high-speed and dynamic environments, demanding algorithms capable of high update frequency and handling highly nonlinear vehicle dynamics at physical limits.
Several approaches have been explored, including model predictive control (MPC) based algorithms for overtaking moving vehicles \cite{buyval2017deriving, zeng2021safety, brudigam2021gaussian, raji2022motion}.
Graph search-based algorithms \cite{rowold2022efficient,stahl2019multilayer}, nonlinear dynamic programming (NLP) \cite{febbo2017moving} and optimization-based local re-planning \cite{he2022autonomous} are also used to solve this problem.
Meanwhile, game theory-based algorithms \cite{liniger2019noncooperative,wang2021game,jung2021game} offer a solution for competing with other racing cars.
Despite their successes, these methods still have limitations.
Research in \cite{febbo2017moving, zeng2021safety,brudigam2021gaussian} focuses solely on overtaking maneuvers in free space environments, neglecting lap time performance.
In \cite{zeng2021safety, brudigam2021gaussian}, the lack of a high-level trajectory planner can lead to deadlocks. 
While algorithms in \cite{buyval2017deriving,stahl2019multilayer, raji2022motion, rowold2022efficient} consider lap time performance, these algorithms can handle only one moving obstacle at a time.
In contrast, the local trajectory planner in \cite{he2022autonomous} optimizes performance along the prediction horizon while competing with multiple cars.
However, its nonsmooth switch between optimizing lap time performance and overtaking other racing cars limits the overall performance.
Meanwhile, high computational burdens of game theory-based algorithms are unsuitable for real-time competition with multiple racing cars due to their complex problem formulations.

Recently, emerging learning-based techniques have been applied to autonomous racing competitions.
For instance, a Bayesian optimization algorithm is used in \cite{jain2020computing} to compute the time-optimal trajectory for a track.
Meanwhile, various learning algorithms-such as deep reinforcement learning (DRL) \cite{perot2017end, remonda2021formula, fuchs2021super, brunnbauer2022latent,zhang2022residual, evans2023high,evans2023safe}, Deep Neural Networks (DNNs) \cite{wadekar2021towards}, Q-learning \cite{gundu2019intermittent}, Imitation Learning (IL) \cite{sun2022benchmark}, Deep Learning (DL) \cite{weiss2020deepracing} and Reinforcement Learning (RL) \cite{spielberg2023learning}-are used to train end-to-end control policies to optimize the vehicle's lap time performance directly. 
In scenarios where the ego vehicle competes with other racing cars, learning-based algorithms also present their capabilities to address this challenge.
A RL-based algorithm \cite{bhargav2021track} is developed to learn overtaking maneuvers in different sections of a specific track offline.
Gaussian process (GP) based strategies are developed in \cite{zhu2023gaussian,benciolini2024active} to predict the future movements of opponents in racing games.
Authors in \cite{song2021autonomous} use RL to train an end-to-end autonomous racing policy and evaluate its lap time performance against other cars in a video game based simulator.
However, learning-based methods often require extensive data and track-specific training.
Overfitting can occur when complex real-world factors, such as partially slippery track surfaces and tire wear, are insufficiently modeled in the video game.
More importantly, while the training process is suitable for video game based simulator, it may not be practical for real-world racing due to safety concerns, particularly the risk of collisions during the training process.

As discussed, existing efforts in planning and control design for autonomous racing cars struggle with improving lap time performance and executing overtake maneuvers simultaneously.
Especially, model-based strategies often suffer from nonsmooth switches between the planner and the controller when the ego vehicle must overtake leading vehicles \cite{he2022autonomous}.
While end-to-end policies obtained through learning-based algorithms seem to avoid such switches, they require extensive training for each specific track, demanding large amounts of data and raising safety concerns in complex real-world racing environments \cite{song2021autonomous, bhargav2021track}.
Furthermore, most existing studies focus on scenarios involving only a few competing vehicles, which fails to capture the complexity of real-world races \cite{song2021autonomous}.
Table~\ref{tab:work_summary} presents a comparative analysis of various strategies and their distinct features within the car-racing scenario.

The iterative linear quadratic regulator (iLQR) \cite{chen2017constrained, howell2022trajectory} has recently gained popularity as a model-based predictive controller offering a solution to the local infeasibility often faced by MPC.
Building on this, and inspiration by iterative learning based control and the recent work~\cite{zeng2023i2lqr}, we propose an autonomous racing algorithm that seamlessly integrates the planner and controller into a unified framework.
The proposed approach improves lap time performance while competing against multiple competitors.
Specifically, it utilizes historical data from previous laps to improve lap time performance while executing overtaking maneuvers on multiple moving racing.
This strategy avoids the nonsmooth switch between the time-optimal controller and overtaking controller, resulting in an improved racing performance.

\begin{table*}
\def\tablename{Table}
\centering
\scriptsize
\begin{tabular}{|c | c| c |c |c |c |c |c |c |c |c |c |c |c|}
\hline
\emph{Approach} & \emph{GP} & \multicolumn{2}{c|}{\emph{DRL}} & \emph{Graph-Search} & \multicolumn{3}{c|}{\emph{Game Theory}} & \multicolumn{6}{c|}{\emph{Model-Based}} \\ \hline
Publication & \cite{hewing2018cautious} &
\cite{fuchs2021super} & 
\cite{song2021autonomous}&
\cite{stahl2019multilayer} & \cite{liniger2019noncooperative} & \cite{wang2021game} & 
\cite{jung2021game} &
\cite{rosolia2017learning} & \cite{kabzan2019learning} & \cite{kapania2020learning} & 
\cite{raji2022motion} & 
\cite{he2022autonomous} & \textbf{Ours} \\ \hline
Lap Timing & \textbf{\checkmark} & \textbf{\checkmark} &\textbf{\checkmark} & \textbf{\checkmark} & \textbf{\checkmark} & $\times$ & \textbf{\checkmark} & \textbf{\checkmark} & \textbf{\checkmark} & \textbf{\checkmark} & \textbf{\checkmark} & \textbf{\checkmark} & \textbf{\checkmark} \\ \hline
Static Obstacle & N/A & N/A &\textbf{\checkmark} & \textbf{\checkmark} & \textbf{\checkmark} & \textbf{\checkmark} & \textbf{\checkmark} & N/A & N/A & N/A & \textbf{\checkmark} & \textbf{\checkmark} & \textbf{\checkmark} \\ \hline
Dynamic Obstacle & N/A & N/A & \textbf{Multi.} & One & One & \textbf{Multi.} & \textbf{Multi.} & N/A & N/A & N/A & N/A & \textbf{Multi.} & \textbf{Multi.} \\ \hline
Update Frequency (Hz) & N/A & N/A & N/A & 15 & \textbf{30} & 2 & N/A & \textbf{20} & \textbf{20} & Offline & \textbf{20} & \textgreater\textbf{25} & \textgreater\textbf{33} \\ \hline
Unified Algorithm & $\times$ & $\times$ & \textbf{\checkmark} & \textbf{\checkmark} & \textbf{\checkmark} & $\times$ & \textbf{\checkmark} & $\times$ & $\times$ & $\times$ & $\times$ & \textbf{\checkmark} & \textbf{\checkmark} \\ \hline
Dynamics Accuracy & N/A & N/A & N/A & \textbf{\checkmark} & \textbf{\checkmark} & $\times$ & \textbf{\checkmark} & \textbf{\checkmark} & \textbf{\checkmark} & \textbf{\checkmark} & \textbf{\checkmark} & \textbf{\checkmark} & \textbf{\checkmark} \\ \hline
\end{tabular}
\normalsize
\caption{A comparison of recent work on autonomous racing. 
``N/A" indicates that an attribute is unclear or not applicable to a given approach. 
``\textbf{Multi.}" signifies that the algorithm was evaluated in an environment containing multiple dynamic obstacles.}
Notably, our proposed method can adeptly handle multiple challenges simultaneously while maintaining a high update frequency.
\label{tab:work_summary}
\end{table*}

\subsection{Contribution}
The contributions of this paper are as follows:
\begin{itemize}
    \item  We present a novel real-time implementable autonomous racing strategy called IteraOptiRacing that is based on Iterative Linear Quadratic Regulator for Iterative Tasks (i2LQR).   
    The proposed strategy unifies the planner and controller into a single algorithm and handles scenarios both with and without multiple surrounding moving vehicles seamlessly.
    \item 
    The proposed strategy avoids the nonsmooth switch between different scenarios seen in previous methods, which improves the performance of the ego racing car.     
    By using historical data of the ego racing car, the proposed algorithm is capable of improving lap time performance across different race tracks in real time. 
    Its unified strategy contributes to the fast overtaking maneuver in the presence of multiple surrounding moving vehicles.
    \item We validate the proposed autonomous racing algorithm using a high-fidelity racing simulator featuring randomly generated moving vehicles. 
    The results demonstrate that our approach enables the ego racing car to overtake multiple moving vehicles more quickly than the previous approach in various racing scenarios while maintaining a steadily low computational complexity.
    This makes it suitable for the fast-changing racing environment.
\end{itemize}


\section{Background}
\label{sec:background}

In this section, we present the vehicle model and introduce the baseline methods used for demonstrating the results.

\subsection{Vehicle Model}
\label{sec:ATV-model}
In this work, a dynamic bicycle model with decoupled Pacejka tire model under Frenet coordinates is used for the algorithm design and numerical simulations \cite{rajamani2011vehicle}.
The nonlinear system dynamics is described as follows,
\begin{equation}
\Dot{x} = f(x, u),
\label{eq:dynamic-model}
\end{equation}
where $x \in \mathbb{R}^n$ and $u \in \mathbb{R}^m$ are the state and input of the vehicle, $f$ represents the nonlinear bicycle dynamics model in~\cite{rajamani2011vehicle}.
The definitions of the system states and inputs are as the following,
\begin{equation}
x=[v_x, v_y, \omega_z, e_\psi, s_c, e_y]^T,~u=[a, \delta]^T,
\end{equation}
where the acceleration at the  vehicle's center of mass of $a$ and steering angle $\delta$ are the system inputs.
$s_c$ is the curvilinear distance travelled along the track's center line, $e_y$ and $e_\psi$ denote the lateral deviation distance and heading angle error between vehicle and center line.
$v_x$, $v_y$ and $\omega_z$ represent the longitudinal velocity, lateral velocity and yaw rate, respectively. Specifically, at time step $t$, we denote the system state and input vectors by $x_t$ and $u_t$.

In Sec.~\ref{sec:algorithm}, to get a relatively accurate discrete-time dynamics model for the algorithm, we use an affine time-varying model from \cite{rosolia2019learning}.
This model directly learns a linear approximation of the system dynamics around $x_t$ using a local linear regressor, and its formulation in IteraOptiRacing is presented in \eqref{eq:i2lqr-dynamics}.

Notice that, the exact nonlinear dynamics for the bicycle model \eqref{eq:dynamic-model} is employed in our high-fidelity simulator using Euler discretization with sampling time \SI{0.001}{\second} (1000 Hz), which is also used for our simulation analysis in Sec.~\ref{sec:results}. 

\subsection{Baseline Autonomous Racing Strategies}

In this subsection, we briefly introduce the three state-of-the-art autonomous racing strategies which are later used as baseline methods in this article.
All the three strategies are based on learning-based MPC control~\cite{rosolia2019learning}.

LMPC uses historical data in an iterative manner to handle autonomous racing tasks.
In such tasks, the autonomous system repeatedly addresses the same challenge across iterations.
Specifically, LMPC optimizes lap time through historical states, inputs and the associated cost-to-go from previous iterations.
This cost-to-go is the time required to complete the lap from the associated state and is determined after each lap's completion.
A tracking controller, such as PID or MPC, is used to collect initial data over the first few laps.

After the initial laps, LMPC is deployed to optimize the ego racing car's behavior based on the collected data.
At each time step, the terminal constraint of a local MPC problem is formulated as a convex set, containing states that drive the car to complete previous laps.
By defining the cost function as a minimum-time problem, this local MPC problem computes a time-optimal open-loop trajectory.
Since the cost function incorporates cost-to-go data from historical states, the ego racing car can complete the lap within a time that is no greater than the time from the same position during previous laps.
Ultimately, the ego racing car converges to time-optimal performance after several laps.
Further details on the LMPC algorithm are available in Appendix~\ref{sec:appendix} and \cite{rosolia2017learning}. 

\subsubsection{LMPC with Local Re-Planning}
\label{sec:lmpc-re-planning}

The LMPC alone cannot handle dynamic obstacles on the race track.
To enable competitive performance among multiple obstacles, as shown in \cite{he2022autonomous}, an autonomous racing strategy that switches between two modes is proposed.
When no surrounding vehicles are present, the LMPC shown in the previous subsection is used to improve the ego vehicle's lap time performance.
During overtaking maneuvers, an optimization-based planner in the MPC framework generates feasible trajectories for passing. 
A low-level MPC controller is used to track them.
Further details on this re-planning strategy are provided in Appendix~\ref{sec:appendix-re-planning} and \cite{he2022autonomous}.

\subsubsection{LMPC with Slacked Target Terminal State}
\label{sec:lmpc-slack-target-state}
In the original LMPC formulation, the open-loop terminal state must lie within the convex hull formed by historical closed-loop states.
However, with other dynamic obstacles on the track, this convex hull may become unreachable by the end of the terminal horizon.
To improve feasibility, authors in \cite{rosolia2021minimum} introduce a modified LMPC algorithm.
In this approach, each individual historical state in the convex hull serves as a target terminal state for the local MPC optimization. 
Slack variables are added to allow for a certain degree of terminal constraint violation.
When the obstacle avoidance constraint is also included, this approach becomes suitable for generating open-loop trajectories in the presence of other racing cars.
The ego racing car then uses the first control input of the feasible trajectory with the minimum cost-to-go.
More details are available in Appendix~\ref{sec:appendix-slack-target} and \cite{rosolia2021minimum}.

\subsubsection{LMPC with Slack on Convex Hull}
\label{sec:lmpc-slack-hull}
To ensure feasibility in the local MPC optimization, the original LMPC algorithm can also be adopted by relaxing the convex hull constraint~\cite{rosolia2021minimum}. 
By adding obstacle avoidance constraints when dynamic obstacles are within the local MPC's prediction horizon, the terminal state of the optimal open-loop trajectory is allowed to lie outside of the convex hull of historical states.
This adjustment enables the ego racing car to overtake other dynamic obstacles while maintaining lap time performance.
Further details can be found in Appendix~\ref{sec:appendix-slack-hull} and \cite{rosolia2021minimum}.
\begin{remark}
    We select these three model-based racing strategies as baselines. Similar to the proposed algorithm, all of these baseline methods are suitable for competing with multiple racing cars while improving lap time performance. 
    In the absence of obstacles, these strategies enhance the ego agent's performance through learning from historical data, while the results in \cite{zeng2023i2lqr} demonstrate that i2LQR achieves the same optimal performance as the LMPC algorithm under obstacle-free conditions, indicating that the performance of the three baselines aligns with that of IteraOptiRacing in such scenarios. In the context of racing, these methods execute multi-vehicle avoidance while optimizing lap time performance.
\end{remark}
\section{Proposed IteraOptiRacing Algorithm}
\label{sec:algorithm}

\begin{figure*}
    \centering
    \includegraphics[width=1\linewidth]{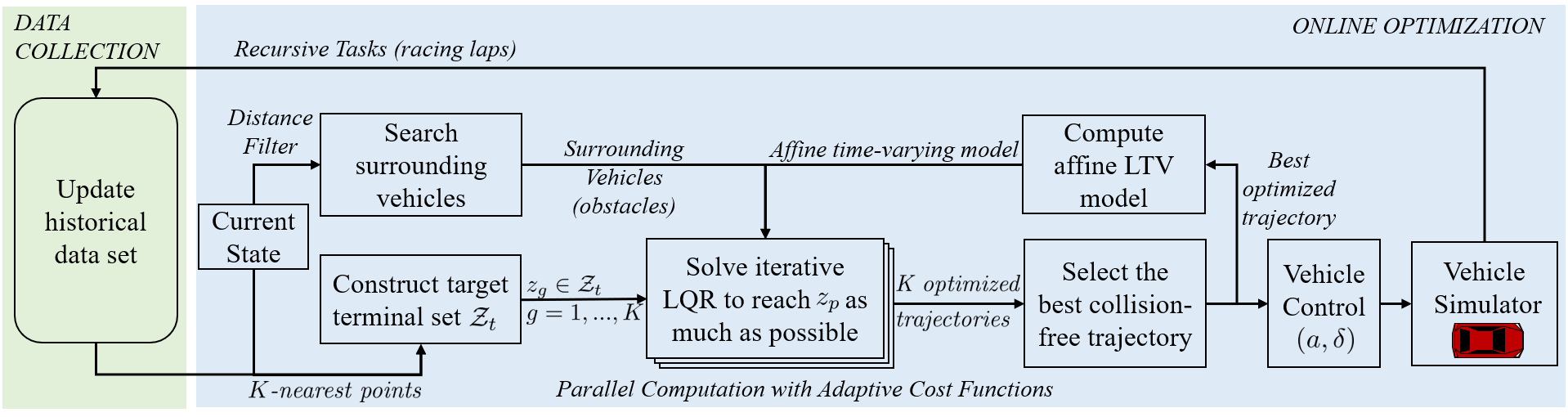}
    \caption{ IteraOptiRacing racing strategy. 
    Initial historical data are collected offline through recursive tasks.
    During the online optimization, a target terminal set $Z_t$ is constructed at each time step using $K$-nearest points from the historical data.
    Together with surrounding vehicles' information, points $z_p$ from this target terminal set will be used to formulate $K$ local iLQR optimization problems, which come with adaptive cost functions and can be solved efficiently through parallel computing.
    The dynamics used for these optimization problems are an affine time-varying model based on the open-loop predicted states and inputs from the last time step.
    Finally, the best collision-free trajectory will be selected among $K$ candidate optimized trajectories.}
    \label{fig:structure_i2lqr}
\end{figure*}

After introducing the vehicle model and baseline racing strategies, we present the proposed autonomous racing planner-controller algorithm in detail. 
This algorithm is designed to assist the ego vehicle in overtaking other moving vehicles while considering the lap time performance.

\subsection{Autonomous Racing Strategy} 
IteraOptiRacing racing strategy is based on the iterative LQR for iterative tasks in dynamic environments (i2LQR) \cite{zeng2023i2lqr}.
Fig.~\ref{fig:structure_i2lqr} illustrates the framework of the overall algorithm.
It comprises the following components: offline data collection and online optimization.

\subsubsection{Data Collection}
During the process of data collection, the algorithm collects the historical data of the ego racing car from the starting point to the finish line for each lap.
This historical data encompasses the ego racing car's past states, inputs, and the cost-to-go associated with each recorded state.
The cost-to-go, same as the one used in the LMPC framework, represents the time required to complete the corresponding lap from the recorded state.
The algorithm stores the historical information into a historical data set $\mathcal{H}$.
To collect initial data, a straightforward tracking controller, such as PID or MPC controller, is employed for the first several laps, which can be conducted offline. Subsequently, the proposed autonomous racing algorithm leverages the recorded data to optimize the performance of the ego racing car while acquiring additional data simultaneously to further enhance its performance.

\subsubsection{Online Optimization}
In online optimization, our strategy not only optimizes lap time based on historical data, but also provides safe trajectories in scenarios with multiple dynamic obstacles to avoid collisions.
At each time step of online optimization, an optimal target state is selected from historical data set $\mathcal{H}$ for iLQR optimization.
To achieve this while considering reachability, we first pick the $K$-nearest neighbors in $\mathcal{H}$ with respect to the current state and construct the target terminal set $\mathcal{Z}_t$.
Next, the local optimization problem iterates through the target terminal set $\mathcal{Z}_t$, selecting each candidate $z_g$ as the target terminal state, with cost-to-go values prioritized in descending order.
Subsequently, an iLQR-based controller will generate the open-loop optimal trajectories for these candidates.
In this process, each $z_g \in \mathcal{Z}_t$ corresponds to an individual local optimization problem, which generates candidate open-loop trajectories for the ego racing car.

In the absence of the surrounding vehicles, our control strategy focuses on optimizing lap time.
After iterating over each candidate $z_g$ in $\mathcal{Z}_t$ and generating trajectories using iLQR, the proposed algorithm selects the optimal candidate $z_g$ based on both time optimality and reachability.
The algorithm evaluates the reachability by determining whether the terminal state of the iLQR-generated trajectory satisfies the tracking error constraint related to $z_g$.
Meanwhile, we evaluate the time optimality by considering the cost-to-go associated with each target terminal state, ensuring the selection of the target terminal states that contribute to minimizing lap time.
By prioritizing reachable target states that minimize the cost-to-go during the selection process, the vehicle is able to reach the finish line in a time no greater than that of the same position in previous laps.
Consequently, the ego vehicle can achieve time-optimal performance after several laps.

When overtaking other vehicles, the proposed algorithm can still optimize lap time while ensuring the safety of the ego racing car.
To achieve this, we convert obstacle avoidance constraints into soft constraints by rewriting them as parts of the cost function.
Similar to the situation with no obstacles, our control strategy iterates through the candidates $z_g \in \mathcal{Z}_t$ in ascending order of the cost-to-go and generates trajectories using iLQR.
However, to facilitate fast and smooth overtaking maneuvers, our algorithm dynamically adjusts the parameters of the local iLQR optimization problem for each candidate $z_g$, taking into account reachability and obstacle avoidance capabilities.
This approach ensures the generation of optimal and safe trajectories that fulfill both objectives, enhancing lap time performance while safely overtaking surrounding vehicles.

In the proposed algorithm, since solving the local optimization problem with different candidate terminal state $z_g$ is independent, the parallel computing technique can be used to speed up the computation process, in which multiple optimized open-loop trajectories can be obtained simultaneously.
After obtaining all the $K$ optimized open-loop trajectories, the best collision-free trajectory is selected by considering the tracking error condition, the potential for collisions with obstacles, and the requirement for time optimality.
The corresponding optimal inputs are then executed by the ego vehicle.

The proposed approach solves the optimization problem using a unified framework, enabling the planning of optimal trajectories for scenarios both with and without surrounding vehicles. This integrated methodology eliminates the nonsmooth transitions between the time-optimal and overtaking controllers observed in previous works, therefore enhancing performance and safety during overtaking maneuvers.
Details about the proposed algorithm are presented in the following subsections.

\subsection{Target Terminal Set Construction}
In this subsection, we introduce the construction of the target terminal set $\mathcal{Z}_t$ in detail using points derived from the historical data set $\mathcal{H}$. 
At time step $t$, $K$-nearest neighbors with respect to the current state $x_t$ are selected to construct the target terminal set using the following criteria:
\begin{subequations}\label{eq:i2lqr-nearest_points}
\begin{align}
    \argmin\limits_{{z}_g} 
    \sum\limits_{g=1}\limits^{K}||{z}_g - &{x}_t||^{2}_{D_z} \\
    \text{s.t.}\quad z_{g} \in \mathcal{H},&~g= 1,...,K,~\text{with all } z_g \text{ distinct}
\end{align}
\end{subequations}
where $\mathcal{H}$ is the historic data set, ${D_z}$ is a diagonal matrix that contains weight coefficients for state variables.
To enhance computational efficiency, only the historical states from the selected previous iterations are used for the selection of $K$-nearest points.
Consequently, these selected points are used as the target terminal states $z_g$ for the local iLQR optimization problems.

\begin{figure}
    \setlength{\abovecaptionskip}{0.35cm}
    \setlength{\belowcaptionskip}{-0.5cm}
    \centering
    \includegraphics[width=1.0\linewidth]{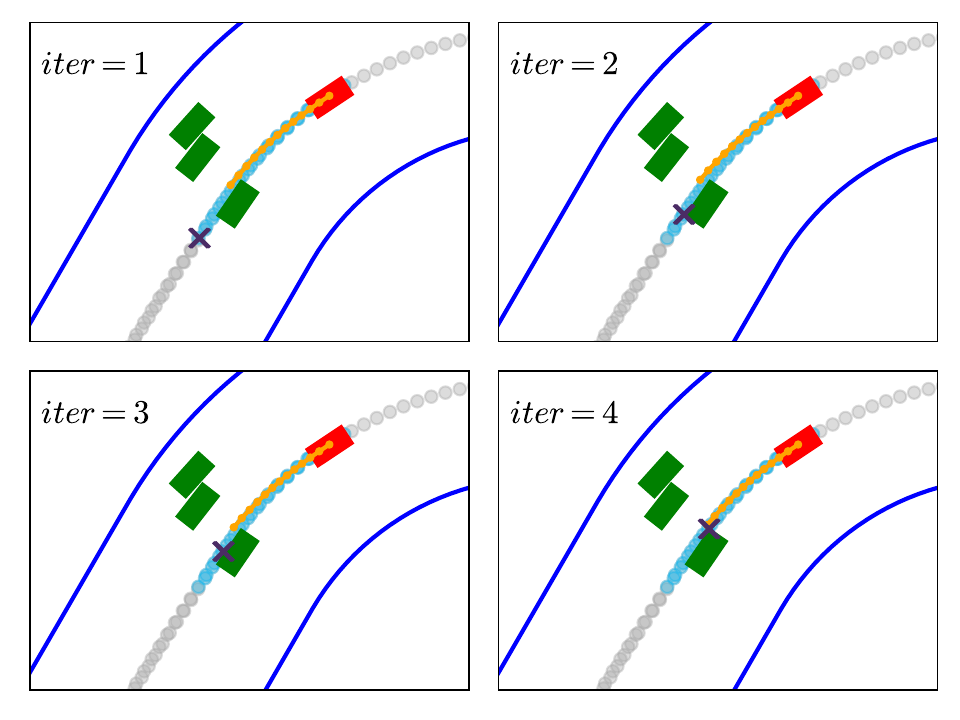}
    \caption{An illustration of the iterative cycle for target terminal state selection where the states on the left assume to have a smaller cost-to-go.
    Ego vehicle at the current time step is marked in red, while dynamic obstacles are marked in green.
    Points in skyblue are the selected nearest points and those in grey are the historical states. 
    Specifically, current state is considered as the guided state defined in \cite{zeng2023i2lqr}, and purple cross is the target terminal state. 
    The orange line is the best open-loop trajectory.
    During different iterations, the algorithm selects varying target terminal states in purple and computes their associated candidate trajectories.}
    \label{fig:iterative_point_selection}
\end{figure}

In this study, without parallel computation, in order to allow the ego vehicle to improve its lap time and reduce the computation time of the algorithm, points $z_g$ with smaller cost-to-go is first used by the local iLQR optimization problem.
This procedure is iteratively performed until the optimal open-loop trajectory aligns with the algorithm's requirements, at which point the optimal control is executed by the ego racing vehicle.
This engineering process enables high-frequency operation even without parallel computation.
A diagram to present this process is shown in Fig.~\ref{fig:iterative_point_selection}.

\subsection{Local iLQR Optimization}
\label{sec:ilqr_problem}
The following constrained finite-time optimal control problem is solved through iLQR for each $z_g\in \mathcal{Z}_t$:

\begin{subequations}
\label{eq:i2lqr}
\begin{align}
    & \min\limits_{{u}_{t:t{+}N{-}1|t}} p({x}_{t+N|t},z_g) + \sum_{k=0}^{N-1} q({x}_{t+k|t}, {u}_{t+k|t})\label{eq:i2lqr-cost}\\
    \text{s.t.} &\ 
    {x}_{t+k+1|t} =  A_{t+k|t} {x}_{t+k|t} + B_{t+k|t}{u}_{t+k|t}\label{eq:i2lqr-dynamics} 
    \\ & +C_{t+k|t}, \ k= 0,...,N{-}1 \nonumber \\ 
    & {x}_{t+k+1|t} \in \mathcal{X},\ {u}_{t+k|t} \in \mathcal{U}, \ k = 0,...,N{-}1 \label{eq:i2lqr-constraint}\\
    & {x}_{t|t} = {x}_t, \label{eq:i2lqr-initial-condition}
\end{align}
\end{subequations}
where \eqref{eq:i2lqr-cost} is the cost of the optimization problem;
$N$ is the horizon length;
\eqref{eq:i2lqr-dynamics} is the affine time-varying dynamics as introduced in Sec.~\ref{sec:ATV-model}, where  $A_{k|t}$, $B_{k|t}$ and $C_{k|t}$ are state-dependent and time-varying matrices;
\eqref{eq:i2lqr-constraint}, \eqref{eq:i2lqr-initial-condition} represent the input/state constraints and initial conditions, respectively.

\subsubsection{Cost Function}
The cost \eqref{eq:i2lqr-cost} includes terminal cost and stage cost. The stage cost is defined as the following:
\begin{equation}\label{eq:stage-cost}
     q({x}_{t+k|t}, {u}_{t+k|t}) = 
     ||{u}_{t+k|t}||^2_R +  ||{u}_{t+k|t} - {u}_{t+k-1|t}||^2_{dR}
\end{equation}
where $k = 1, ..., N{-}1$, $R \in \mathbb{R}^{2 \times 2}$ and $dR \in \mathbb{R}^{2 \times 2}$ are adaptive stage weights.
The terminal cost introduces the difference between the open-loop predicted terminal state and the target terminal state $z_g$ from target terminal set $Z_t$ in the quadratic form:
\begin{equation}\label{eq:terminal-cost}
\begin{aligned}
p({x}_{t+N|t},z_g) =  ||x_{t+N|t} - z_{g}||^2_{Q_N} 
\end{aligned}
\end{equation}
where $Q_N \in \mathbb{R}^{6 \times 6}$ is an adaptive terminal weight.

\subsubsection{Reformulating State and Input Constraints as Cost Terms}
To solve the above optimization problem \eqref{eq:i2lqr} through iLQR, we transform the constraints on states and inputs to part of the new cost function through the exponential function used in \cite{chen2017constrained},
\begin{equation}\label{eq:exp-barrier-func}
    c = q_1 \exp(q_2 f)
\end{equation}
where $f \leq 0$ represents the constraint, $q_1$ and $q_2$ are the tuning weights.
Specifically, for the obstacle avoidance constraint, the following constraint is formulated:
\begin{equation}\label{eq:obs-ellipse}
\begin{aligned}
    1 - ||x_{t+k|t} - x_{p,t+k|t}||^2_P < 0
\end{aligned}
\end{equation}
which approximately treats the vehicle as an ellipse at step $k$ of the prediction horizon.
$x_{t+k|t}$ is the ego vehicle's predicted state at step $k$ in the prediction horizon; 
$x_{p,t+k|t}$ is the $p$-th surrounding vehicle's predicted state at step $k$. 
$P \in \mathbb{R}^{6 \times 6}$ represents the weight matrix as follows,
\begin{equation}
    P = diag(0, 0, 0, 0, (l+v_{{x}_{t+k|t}} t_{\text{safe}}+s_\text{safe})^{-2}, (d+s_\text{safe})^{-2}), \label{eq:matrix-P}
\end{equation}
where $l$ and $d$ are the vehicle length and width, respectively.
$v_{x}$ is the ego vehicle's velocity at step $k$ of the prediction horizon, $t_\text{safe}$ is the safe head way and $s_\text{safe}$ is a safety margin.
This condition is included as a part of the cost function.

\begin{remark}
    To ensure the safety of the system and generate collision-free trajectories, we implement an iterative process for each $z_g$, in which our strategy continuously adjusts the weights of the cost and constraints, solving the optimization problem in \eqref{eq:i2lqr} based on the updated weights.
    At each iteration, a trajectory is first generated using iLQR along with the weights from the previous iteration.
    This trajectory is then evaluated to determine whether it satisfies the following condition:
    \begin{equation}\label{eq:safe_boundary}
    \begin{aligned}
    (s_{c_{t+k}} - s_{c,p_{t+k}})^2 + (e_{y_{t+k}} - e_{y,p_{t+k}})^2 - l^2 - d^2 > 0,
    \end{aligned}
    \end{equation}
    where $k = 1,...,N$, $s_{c_{t+k}}$ and $e_{y_{t+k}}$ are the ego vehicle's open-loop traveling distance and lateral deviation at time step $t+k$ of the prediction horizon in the iLQR optimization, respectively;
    $s_{c_,p_{t+k}}$ and $e_{y,p_{t+k}}$ are the $p$-th surrounding vehicle’s predicted traveling distance and lateral deviation at time step $t+k$ of the prediction horizon, respectively.
    If the above condition \eqref{eq:safe_boundary} is not satisfied, our strategy compromises reachability to ensure collision-free trajectories in the next iteration.
    Specifically, the tracking weight $Q_N$ in \eqref{eq:terminal-cost} is adjusted to $Q_N/m_{Q_N}$. Similarly, $R$ and $dR$ in \eqref{eq:stage-cost} are updated to $R/m_{R}$ and $dR/m_{dR}$.
    Furthermore, the weight of the obstacle avoidance constraint is increased. Specifically, the parameter $q_2$ in the corresponding exponential barrier function \eqref{eq:exp-barrier-func} is scaled by a factor of $q_2(1 + m_{q_2})$. This adjustment ensures that the iLQR algorithm converges towards satisfying the avoidance constraint during the backward pass.
    The above hyperparameters ($m_{Q_N}>1$, $m_{R}>1$, $m_{dR}>1$, $0<m_{q_2}<1$) can be tuned.
    This iterative process repeats until the trajectory generated by iLQR satisfies the safety conditions in \eqref{eq:safe_boundary} or the maximum iteration number is reached.
\end{remark}

\subsection{Best Collision-Free Trajectory Selection}
In this section, we introduce how to select the best collision-free trajectory in multi-vehicle competition scenarios.
For each target terminal state $z_{g} \in \mathcal{Z}_{t}$, the optimization problem \eqref{eq:i2lqr} is solved by iLQR optimization mentioned in Sec.~\ref{sec:ilqr_problem}.
Therefore, we can get the optimal open-loop state and control sequences associated with each target terminal state $z_g$.
To determine the best optimized trajectory, we apply the following two criteria: (a) whether the open-loop trajectory will collide with any surrounding vehicles at the next time step, and (b) whether the open-loop terminal point $x_{t+N|t}$ can reach the target terminal state $z_g$.
Additionally, we adjust the parameters to assess the reachability based on whether the scenario involves overtaking.
For overtaking scenarios, the following criteria is employed to ascertain whether the $i$-th surrounding vehicle is within the ego vehicle's overtaking range:
\begin{equation}
-\epsilon l \leq s_{c, p} - s_{c} \leq \epsilon l + \gamma |v_{x} - v_{x, i}|
\label{eq:overtaking_criteria}
\end{equation}
where $s_{c}$ and $s_{c, p}$ are the ego vehicle's and the $p$-th surrounding vehicle's traveling distance;
$v_{x}$ and $v_{x, p}$ denote their longitudinal speeds; $l$ is the length of the vehicle; and $\epsilon$ and $\gamma$ are tunable parameters for the safety margin and prediction ratios, respectively.
We first use the safe boundary \eqref{eq:safe_boundary} to check whether the generated trajectory is collision-free with the surrounding vehicles.
If it is collision-free, then we check its reachability.
Two metrics are used for this purpose: the tracking ratio and the convergence ratio.
The tracking ratio is evaluated using $\varepsilon_1$ and $\varepsilon_2$, which checks whether the terminal state of the open-loop trajectory is sufficiently close to the target terminal state $z_g$:
\begin{equation}\label{eq:track_error}
\begin{aligned}
     ||{x}_{t+N|t} - {z}_{g}||^2 < \varepsilon
\end{aligned}
\end{equation}
where $\varepsilon = \varepsilon_1$ when no other vehicle exists and $\varepsilon = \varepsilon_2$ when competing with other racing cars.
If the above condition is not satisfied, we further check whether the optimal solution from the previous iteration, $i-1$, and the current iteration, $i$, meet the convergence criterion:
\begin{equation}\label{eq:convergence_ratio}
\begin{aligned}
     ||{x}_{i-1, t+N|t} - {x}_{i, t+N|t}||^2/||{x}_{i-1, t+N|t}||^2 < \psi
\end{aligned}
\end{equation}
where $\psi = \psi_1$ when no other vehicle exists and $\psi = \psi_2$ when competing with other racing cars.
Finally, among all trajectories that pass all above checks, the optimal open-loop trajectory of $z_{g}$ associated with the minimal cost-to-go $h(z_g)$ is selected as the best optimized trajectory.
The optimal input $u^{*}_{t|t}$ associated with this trajectory is applied to the ego vehicle.
\begin{remark}
    To ensure optimal lap time performance and improve the success rate of overtaking maneuvers, two distinct sets of parameters, $\varepsilon_1$, $\varepsilon_2$, and similarly, $\psi_1$, $\psi_2$, are employed. When no surrounding vehicles are present, smaller values for the tracking ratio, $\varepsilon_1$, and the convergence criterion, $\psi_1$, are used to facilitate faster trajectory convergence and enhanced lap time performance. Conversely, when the ego vehicle is competing with other vehicles, larger values, $\varepsilon_2$ and $\psi_2$, are applied to ensure the generation of safe, collision-free overtaking trajectories.
\end{remark}
\begin{remark}
    Compared with the i2LQR algorithm proposed in \cite{zeng2023i2lqr}, the IteraOptiRacing strategy eliminates the need for selecting the optimal target terminal state in an iterative manner during nearest points selection, thereby reducing computational time in real-time applications.
    Additionally, when solving the local iLQR optimization problem for each candidate $z_g$ to generate collision-free trajectories, our algorithm dynamically adjusts the parameters of each cost term to enhance obstacle avoidance capabilities.
    Finally, in the Best Collision-Free Trajectory Selection process, the IteraOptiRacing strategy incorporates both the tracking ratio in \eqref{eq:track_error} and the error convergence ratio in \eqref{eq:convergence_ratio} as selection criteria. This ensures the generation of safe, collision-free overtaking trajectories, even in scenarios with multiple surrounding vehicles.
\end{remark}
\section{Results}\label{sec:results}
After introducing the proposed autonomous racing algorithm, IteraOptiRacing, we now proceed to validate its performance through numerical simulations.
In this section, we describe the simulation setup and present the corresponding results.

\subsection{Simulation Setup}
In all simulations, the vehicles are 1:10 scale RC cars, each measuring $l = \SI{0.4}{\meter}$ in length and $d = \SI{0.2}{\meter}$ in width.
The ego vehicle's speed ranges from $\SI{0}{\meter/\second}$ to $\SI{1.5}{\meter/\second}$, with maximum acceleration and deceleration of $\SI{1}{\meter/\second^2}$.
The speed ranges of the surrounding vehicles vary depending on the specific setup, but their maximum acceleration and deceleration match those of the ego vehicle. Each surrounding vehicle is controlled by a PID controller, which guides it based on randomly assigned target velocities and lateral deviations.
This randomized configuration emulates the diverse and unpredictable behaviors of surrounding vehicles, creating a realistic and challenging environment. By incorporating such variability, the setup rigorously evaluates the robustness and efficacy of our approach through comprehensive randomized stress testing.
The following sections will provide a detailed explanation of the random target velocity and lateral deviation settings.

\subsubsection{Randomization of Target Lateral Deviation for Surrounding Vehicles}
\label{sec:randomize-surrounding-vehicles}

For all random scenarios mentioned later in this section, surrounding vehicles' target lateral deviation, denoted as $\mathbf{d}(t)$, is a time-dependent function.
It is randomized by combining two components: a low-frequency component $\mathbf{d}_{\text{low}}(t)$ and a high-frequency component $\mathbf{d}_{\text{high}}(t)$.
The total target lateral deviation at any time $t$ is given by:
\begin{equation}\label{eq:total-target-lat-deviation}
\begin{aligned}
\mathbf{d}(t) = \mathbf{d}_{\text{low}}(t) + \mathbf{d}_{\text{high}}(t)
\end{aligned}
\end{equation}
The low-frequency lateral deviation component $\mathbf{d}_{\text{low}}(t)$ is updated every 12 time steps. 
Initially, $\mathbf{d}_{\text{low}}(t)$ is sampled from a uniform distribution:
$\mathbf{d}_{\text{low}}(t=0) \sim \mathcal{U}\left(\SI{-0.7}{\meter}, \SI{0.7}{\meter}\right)$.
Afterward, it is adjusted every 12 time steps by adding a random variation $\Delta \mathbf{d}_{\text{low}}(t)$, which is sampled from $\mathcal{U} (\SI{-0.2}{\meter}, \SI{0.2}{\meter})$.
Between updates, the value of $\mathbf{d}_{\text{low}}(t)$ remains constant.

The high-frequency component $\mathbf{d}_{\text{high}}(t)$ follows a similar structure but with more frequent updates.
It is updated every 6 time steps and introduces finer variations in the target lateral deviation.
At time $t=0$, the high-frequency component is initialized with a value drawn from $\mathbf{d}_{\text{high}}(t=0) \sim \mathcal{U}\left(\SI{-0.15}{\meter}, \SI{0.15}{\meter}\right)$.
Subsequently, at every 6 time steps, it is adjusted by a random variation $\Delta \mathbf{d}_{\text{high}}(t)$, sampled from: $\mathcal{U}\left(-0.1\, \si{\meter}, 0.1\, \si{\meter}\right)$.
This component ensures that smaller, more frequent variations are introduced into the lateral deviation.

This combination of randomized values ensures a dynamic range of lateral deviations, thereby guaranteeing the diversity of maneuvers exhibited by surrounding vehicles in the randomized scenarios.

\subsubsection{Randomization of Target Velocity for Surrounding Vehicles}

In several batches of tests, we introduce randomness to the target velocities of surrounding vehicles, categorizing these velocities into three distinct intervals: $V_1 = [\SI{0.2}{\meter/\second}, \SI{0.4}{\meter/\second}]$,
$V_2 = [\SI{0.4}{\meter/\second}, \SI{0.6}{\meter/\second}]$ and $V_3 = [\SI{0.6}{\meter/\second}, \SI{0.8}{\meter/\second}]$.
The target velocity $\mathbf{v}(t)$ is updated every 12 time steps in a piecewise manner as follows:
\begin{equation}
    \mathbf{v}(t) = 
    \begin{cases} 
    \mathbf{v}(t - 1), & t \neq 12k \\
    \mathcal{U}(V_i), & t = 12k, \, k \in \mathbb{Z}^+
    \end{cases}
\end{equation}
where $i \in \{1, 2, 3\}$ and $\mathcal{U}(V_i)$ denotes that the velocity is randomly drawn from the corresponding interval at every 12 time steps update.

The track's width is set to \SI{2}{\meter}.
The horizon length for the proposed autonomous racing strategy is $N=12$.
For a fair comparison in this section, we maintain consistency in the controller settings, specifically by using the same prediction horizon and number of nearest points selected for all baseline methods, including LMPC with local re-planning, LMPC with Target Slack, and LMPC with Slack on Convex Hull.
Furthermore, all methods employ two iterations during the optimization process.
In addition, for the setup of random scenarios, all random trajectories of surrounding vehicles in the subsequent subsections' tests are pre-generated and stored in advance, ensuring that the trajectories are deterministic for each test.
The discretization time $\Delta t$ is set to \SI{0.1}{\second}.
The chosen values of hyperparameters discussed in Sec.~\ref{sec:algorithm} used for numerical simulations can be found in Table~\ref{tab:hyperparameters-values}.
\begin{table}[]
\def\tablename{Table}
\centering
\begin{tabular}{|l|l|}
\hline
Description & Notation \& Value \\ \hline
Number of nearest points selected & $K = 32$  \\
Adaptive ratio of terminal weight & $m_{Q_N} = 20$ \\ 
Adaptive ratios of stage weights & $m_{R}, m_{dR} = 5, 1.1$ \\
Adaptive ratios of constraint weights & $m_{q_2} = 0.1$ \\
Safety-margin ratio in \eqref{eq:overtaking_criteria} & $\epsilon = 5$ \\ 
Prediction ratio in \eqref{eq:overtaking_criteria} & $\gamma = 2$ \\
Safety margin in \eqref{eq:matrix-P} & $s_\text{safe} = \SI{0.1}{\meter}$ \\
Safe time in \eqref{eq:matrix-P} & $t_\text{safe} = \SI{2}{\second}$ \\
Tracking ratio without obstacles in \eqref{eq:track_error} & $\varepsilon_1 = 0.4$ \\ 
Tracking ratio with obstacles in \eqref{eq:track_error} & $\varepsilon_2 = 1.0$ \\
Convergence ratio without obstacles in \eqref{eq:convergence_ratio} & $\psi_1 =  0.0$ \\ 
Convergence ratio with obstacles in \eqref{eq:convergence_ratio} & $\psi_2 = 0.03$ \\ \hline
\end{tabular}
\normalsize
\caption{Values of hyperparameters for numerical simulations}
\label{tab:hyperparameters-values}
\end{table}

The simulations are implemented in Python, with the optimization modeled in CasADi and solved using IPOPT on Ubuntu 18.04.6 LTS, utilizing a CPU i7-8700 processor with a base clock rate of 3.20 GHz.
Notice that this work does not consider any interaction between the ego vehicle and other surrounding vehicles, such as Stackelberg games~\cite{wang2019game}.

\subsection{Performance Analysis of IteraOptiRacing in Racing Scenarios}
\begin{figure}
    \centering
    \includegraphics[width=1\linewidth]{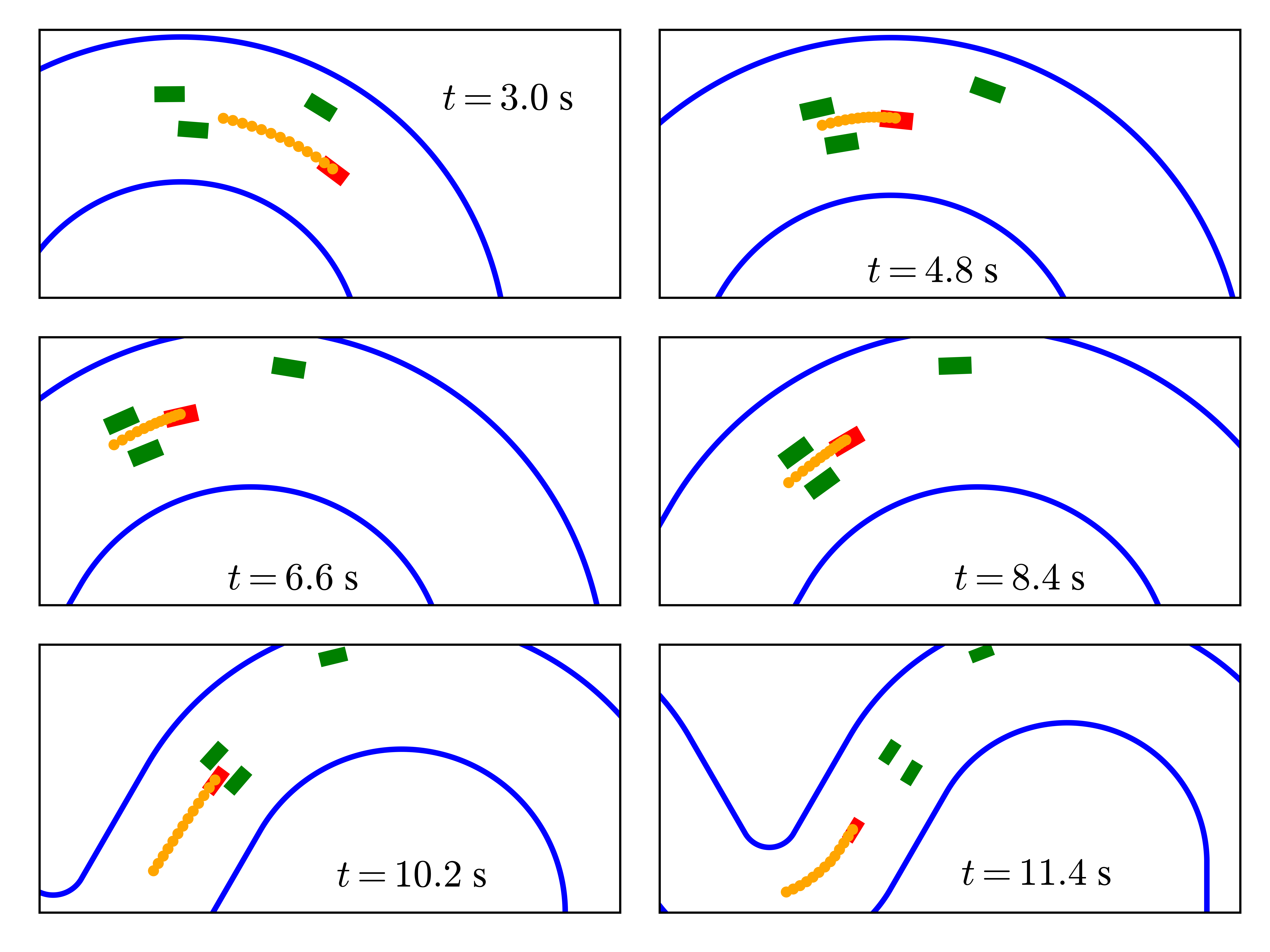}
    \caption{Snapshots from simulation of the overtaking behavior using IteraOptiRacing. The red rectangle represents the ego vehicle, and the green rectangle represents the moving obstacles. The orange line shows the open-loop predicted trajectory of IteraOptiRacing. 
    The track's boundary is marked with solid blue line.}
    \label{fig:result_i2lqr}
\end{figure}

In this subsection, we analyze the performance of our proposed algorithm in racing scenarios, where the ego vehicle is competing with multiple surrounding moving vehicles.
The snapshots shown in Fig.~\ref{fig:result_i2lqr} demonstrate the overtaking capability of the ego vehicle using the proposed algorithm when it competes with three surrounding vehicles.
From the snapshots, it is evident that the ego vehicle smoothly overtakes through a narrow space between the surrounding vehicles.

To further validate our algorithm, we conduct a comparative analysis with the baseline racing strategies: LMPC with local re-planning, LMPC with Target Slack, and LMPC with Slack on Convex Hull in the following subsections.
In the following subsections, we denote LMPC with local re-planning as Baseline 1, LMPC with Slack on the Convex Hull as Baseline 2, and LMPC with Target Slack as Baseline 3.
This evaluation aims to assess the performance and limitations of our proposed autonomous racing algorithm in complex racing environments.

\subsection{Distribution Analysis in Random Overtaking Tests with Different Speed Range}
In this section, we compare the time taken by our approach and baseline racing strategies to overtake the leading vehicles traveling at different speed ranges.
The track utilized in this section is in an M-shaped configuration, with a total length of $\SI{51} {\meter}$, containing a total of 9 surrounding vehicles competing with the ego vehicle.
In each racing scenario, the ego vehicle starts from the origin of the track, while the surrounding vehicles start from a random curvilinear distance within the range of $\SI{5}{\meter} \leq s_{c,p} \leq \SI{40}{\meter}$.
The surrounding vehicles' speeds and lateral deviations are also randomly generated in specific segments introduced in Sec.~\ref{sec:randomize-surrounding-vehicles}. 
To evaluate and compare the performance of our racing strategy with respect to the baseline strategies, we conducted 100 randomly generated tests for each speed range, resulting in  300 tests in total across all speed ranges.

\begin{figure}
    \centering
    \includegraphics[width=1\linewidth]{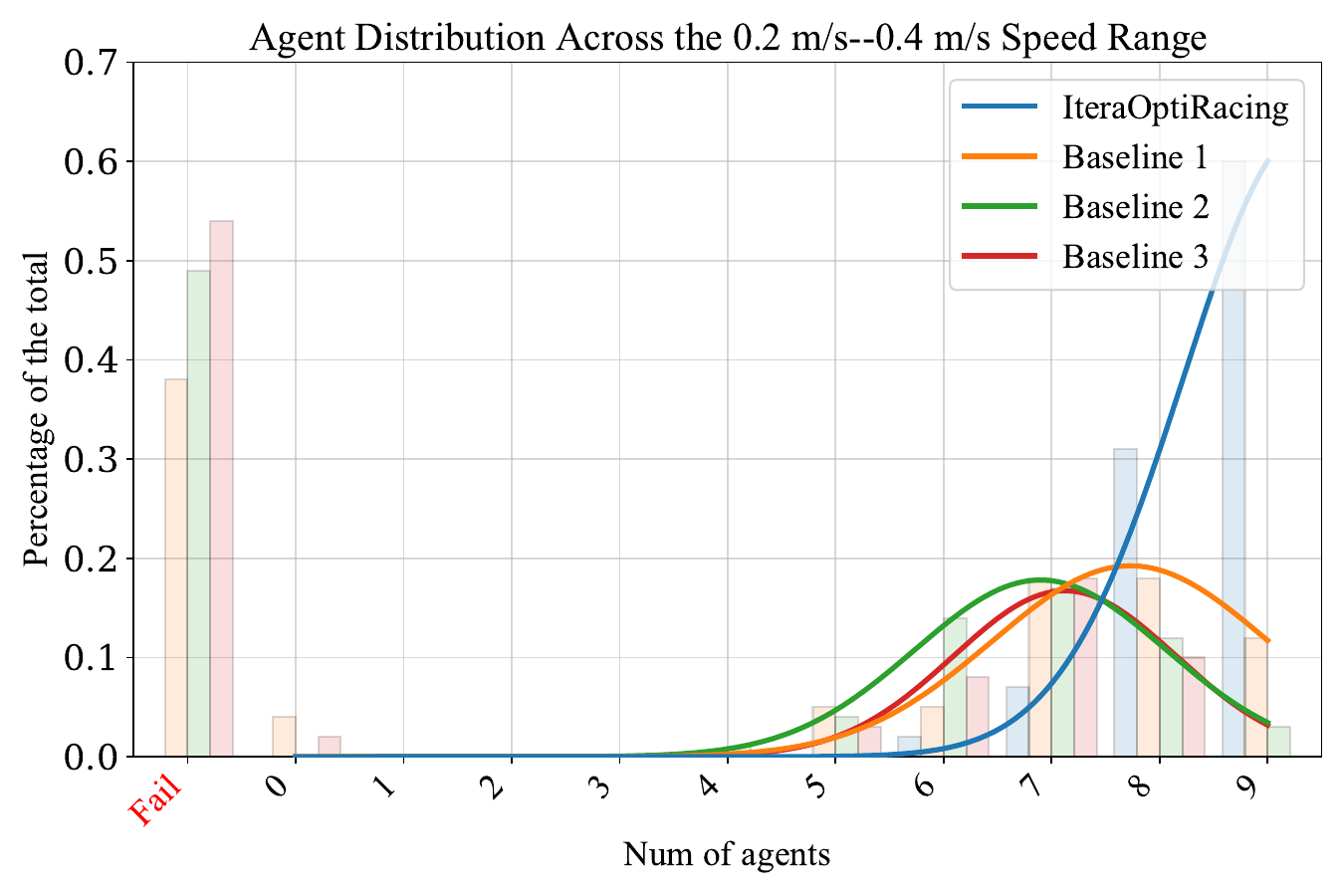}
    \caption{Distribution analysis of overtaking performance across the \SI{0.2}{\meter/\second}-\SI{0.4}{\meter/\second} speed range with 9 obstacle vehicles. The bar chart represents the statistical results of the number of overtaken vehicles across 100 tests, while the curve is obtained by fitting a Gaussian distribution to the data from the bar chart. The success rate of our approach in overtaking all vehicles is significantly higher than that of other baseline methods.}
    \label{fig:distribution-result-spd-range-2to4}
\end{figure}

\begin{figure}
    \centering
    \includegraphics[width=1\linewidth]{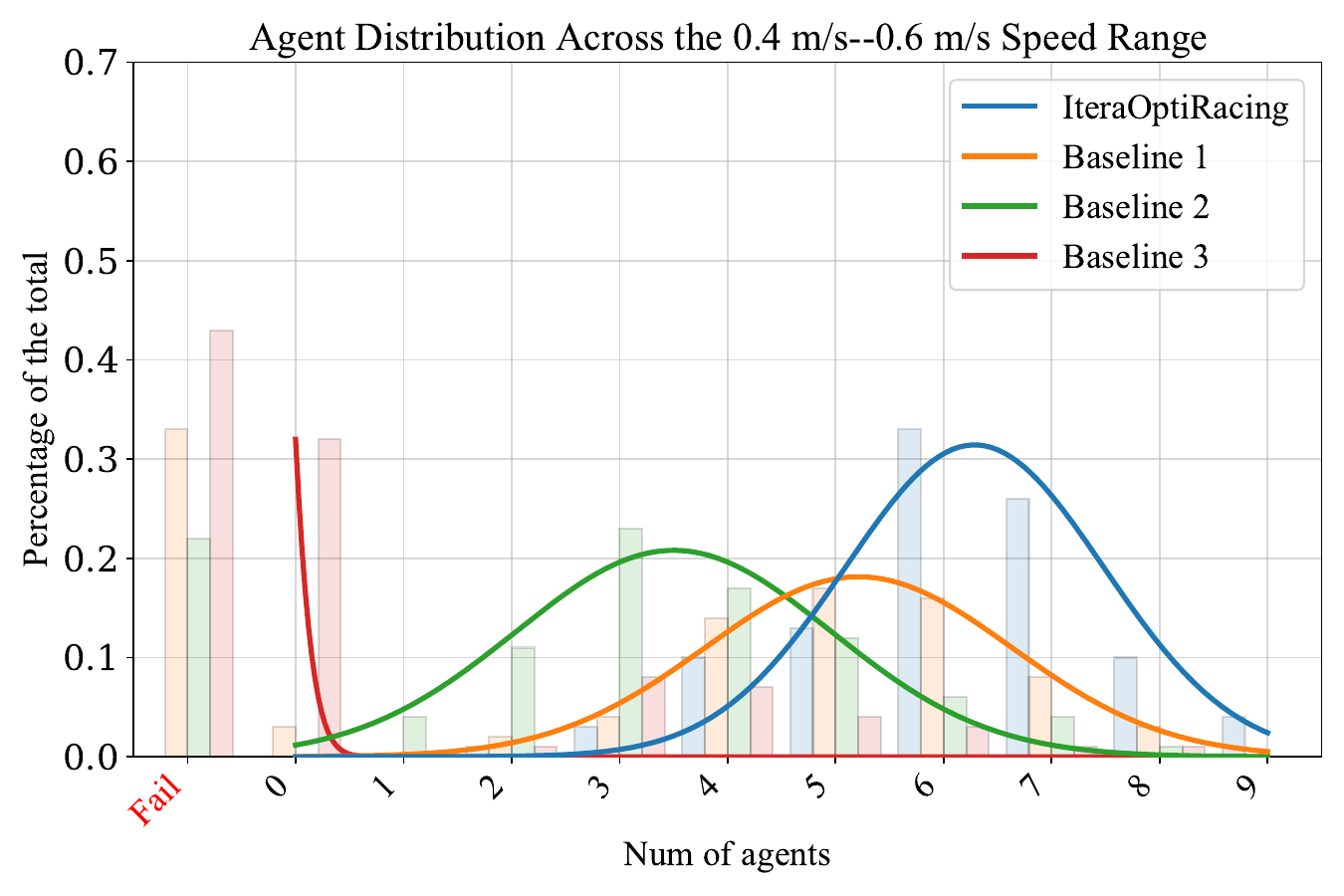}
    \caption{Distribution analysis of overtaking performance across the \SI{0.4}{\meter/\second}-\SI{0.6}{\meter/\second} speed range with 9 obstacle vehicles. Our proposed method overtakes more racing cars on average compared with other baseline methods, with a peak around 6-7 racing cars.}
    \label{fig:distribution-result-spd-range-4to6}
\end{figure}

\begin{figure}
    \centering
    \includegraphics[width=1\linewidth]{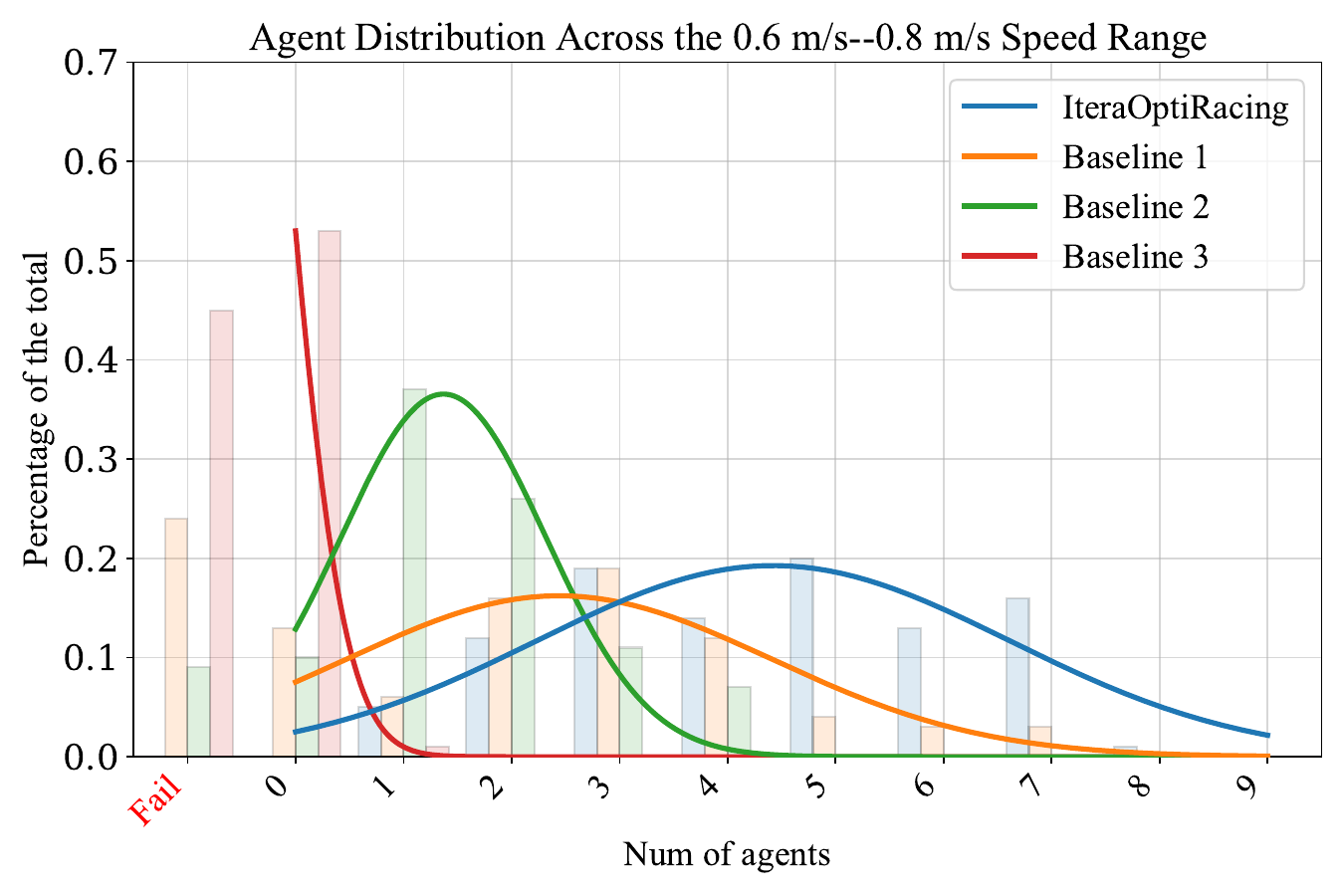}
    \caption{Distribution analysis of overtaking performance across the \SI{0.6}{\meter/\second}-\SI{0.8}{\meter/\second} speed range with 9 obstacle vehicles.}
    \label{fig:distribution-result-spd-range-6to8}
\end{figure}

Figs.~\ref{fig:distribution-result-spd-range-2to4}, \ref{fig:distribution-result-spd-range-4to6} and \ref{fig:distribution-result-spd-range-6to8} respectively show the distribution of the number of obstacles successfully overtaken by each controller across 100 tests in scenarios with 9 obstacle vehicles within specific speed ranges.
The distribution for each test is measured either upon the ego vehicle completing a full lap or reaching the maximum simulation time limit of $\SI{110} {\second}$.
In all three figures, the distribution of the number of obstacle vehicles overtaken by our approach is notably more concentrated on the right side compared to the other three methods.
This indicates that our method is more effective in overtaking within the specified speed range than the alternative approaches, allowing for the overtaking of more obstacle vehicles in less time, which represents smoother overtaking maneuvers.
In addition, it can be observed that, compared to the other three methods, IteraOptiRacing does not experience any failures due to conflicts with boundary constraints, control constraints, or obstacle constraints across all the random tests.
The smooth maneuvers that our approach provides are essential to avoid abrupt movements that could lead to failure, ensuring not only effective performance but also the overall safety of the system.

\subsection{Success Rate in Random Overtaking Tests Across Different Tracks}
\label{sec:sucess-rate-across-different-tracks}
To further compare the proposed autonomous racing strategy with the other three approaches, we not only analyze their overall result distributions but also explore their performance in each individual test.
In this section, we utilize three track shapes: L-shape, M-shape, and ellipse, each approximately $\SI{51}{\meter}$ in length.
For each batch of tests, the ego vehicle is simulated alongside 9 surrounding vehicles whose speeds are randomized within a specified range, as detailed in Sec.~\ref{sec:randomize-surrounding-vehicles}.
Similarly, the initial positions of the obstacle vehicles are randomized within the range of $\SI{5}{\meter} \leq s_{c,i} \leq \SI{40}{\meter}$.
We define ``success" as overtaking all surrounding vehicles within the track by the time the ego vehicle reaches the finish line.
Then, we examine whether the proposed autonomous racing algorithm with i2LQR consistently outperforms the other three methods in every case across different scenarios.
Table~\ref{tab:overtaking-rate-i2lqr-lmpc} presents a comparison of the overtaking success and failure rates for the ego vehicle after completing one lap.
As expected, as the speed range increases with faster surrounding vehicles, there are fewer opportunities and less time available for the ego vehicle to pass safely.
Although the success rate drops for all the algorithms, our proposed autonomous racing strategy still shows its advantages for more complicated racing scenarios.
In fact, for all groups of simulations, there is no instance in which either LMPC with local replanning, LMPC with Target Slack, or LMPC with Slack on Convex Hull successfully overtakes all leading vehicles, while our approach fails to do so.
This highlights that the proposed racing strategy consistently outperforms previous approaches in every test scenario, even as complexity increases.
\begin{table}
\def\tablename{Table}
\centering
\begin{tabular}{|c|cccccc|}
\hline
\multirow{2}{*}{\begin{tabular}[c]{@{}c@{}}Random Speed \\ Range $(\SI{}{\meter/\second})$\end{tabular}} & \multicolumn{6}{c|}{\begin{tabular}[c]{@{}c@{}} Success Rates\\ \hline $^a$: IteraOptiRacing and at least one baseline strat\\-egy succeeds, $^b$: only IteraOptiRacing succeeds \\ $^c$: only baseline strategies succeed, $^d$: all fail \end{tabular}} \\ \cline{2-7} 
 & \multicolumn{2}{c|}{L shape} & \multicolumn{2}{c|}{Ellipse} & \multicolumn{2}{c|}{M shape} \\ \hline
\multirow{2}{*}{$[\SI{0.2}{}, \SI{0.4}{}]$} & \multicolumn{1}{c}{24\%$^{a}$} & \multicolumn{1}{c|}{\bf{62\%}$^{b}$} & \multicolumn{1}{c}{17\%} & \multicolumn{1}{c|}{\bf{26\%}} & \multicolumn{1}{c}{12\%} & \bf{49\%} \\ 
 & \multicolumn{1}{c}{\bf{0\%}$^{c}$} & \multicolumn{1}{c|}{14\%$^{d}$} & \multicolumn{1}{c}{\bf{0\%}} & \multicolumn{1}{c|}{57\%} & \multicolumn{1}{c}{\bf{0\%}} & 39\% \\ \hline
\multirow{2}{*}{$[\SI{0.4}{}, \SI{0.6}{}]$} & \multicolumn{1}{c}{1\%} & \multicolumn{1}{c|}{\bf{4\%}} & \multicolumn{1}{c}{1\%} & \multicolumn{1}{c|}{\bf{2\%}} & \multicolumn{1}{c}{0\%} & \bf{4\%}  \\
 & \multicolumn{1}{c}{\bf{0\%}} & \multicolumn{1}{c|}{95\%} & \multicolumn{1}{c}{\bf{0\%}} & \multicolumn{1}{c|}{97\%} & \multicolumn{1}{c}{\bf{0\%}} & 96\% \\ \hline
 \multirow{2}{*}{$[\SI{0.6}{}, \SI{0.8}{}]$} & \multicolumn{1}{c}{0\%} & \multicolumn{1}{c|}{\bf{1\%}} & \multicolumn{1}{c}{0\%} & \multicolumn{1}{c|}{\bf{1\%}} & \multicolumn{1}{c}{0\%} & \bf{0\%} \\ 
 & \multicolumn{1}{c}{\bf{0\%}} & \multicolumn{1}{c|}{99\%} & \multicolumn{1}{c}{\bf{0\%}} & \multicolumn{1}{c|}{99\%} & \multicolumn{1}{c}{\bf{0\%}} & 100\% \\ \hline
\end{tabular}
\normalsize
\caption{Success rates for IteraOptiRacing compared with baseline strategies in different scenarios after one lap. For each batch of tests, 100 randomized cases are simulated where statistical results about whether overtake maneuvers are successful are classified into four categories marked with following superscripts: (a) IteraOptiRacing succeeds and at least one baseline strategy succeeds (b) only IteraOptiRacing succeed (c) only baseline strategies succeed (d) all fail. These superscripts apply to each block of the table.  Analysis of the statistical results, which include various tracks and different randomized speeds, reveals that there is no case in which the baseline strategies outperform ours. In contrast, in all other scenarios, our approach consistently surpasses the baseline strategies, e.g., percentage value in (c) is always zero, while our proposed strategy always outperforms baseline strategies in various cases. e.g., percentage value in (b) is always bigger than or equal to zero.}
\label{tab:overtaking-rate-i2lqr-lmpc}
\end{table}

\begin{table}
\def\tablename{Table}
    \centering
    \begin{tabular}{|c|c|c|c|c|}
    \hline
    Approach & IteraOptiRacing & \thead{Baseline 1} 
    & \thead{Baseline 2} & \thead{Baseline 3} \\ \hline 
    mean [s] & 0.036 & 0.309 & 0.456 & 0.485 \\ \hline
\end{tabular}
\normalsize
\caption{Average computational time when overtaking the surrounding vehicles for randomized tests described in the set up in Section~\ref{sec:sucess-rate-across-different-tracks}. The results demonstrate that our proposed strategy outperforms the other approaches in terms of computational efficiency during overtaking.
}
\label{tab:computational-time}
\end{table}

In order to compare the solving time of our method and other three methods, we collected the computational time for overtaking in each case across different tracks.
In each case, the overtaking phase of the ego vehicle is identified using the criteria outlined in \eqref{eq:overtaking_criteria}, and the computational time is measured specifically during this phase.
Table~\ref{tab:computational-time} presents the categorized results for each controller, showing the average solving time.
The data demonstrates that ours requires less computational time than the other methods.

\section{Conclusion}\label{sec:conclusion}
We present a unified approach to a planning and control algorithm in autonomous racing called IteraOptiRacing.
We demonstrate the robustness and performance of the proposed strategy through numerical simulations, where surrounding vehicles start from random positions and move with randomized speeds and lateral deviations.
The results show that the proposed algorithm outperforms the state-of-the-art methods in different racing scenarios.
Moreover, future work will address the interactions between the ego vehicle and other competitors, considering how the strategic decisions of both the ego vehicle and other competitors may mutually influence their trajectories and overall performance.

\balance
\bibliographystyle{IEEEtran}
\bibliography{main}{}

\balance

\appendix\label{sec:appendix}

\subsection{Learning-based MPC}\label{sec:appendix-lmpc}
By using the historical data from previous iterations, the LMPC algorithm drives the system from the starting state $x_S$ to the terminal set $\mathcal{X}_F$ in the least possible time.
Mathematically, the following minimum-time optimal control problem is solved:
\begin{subequations}\label{eq:lmpc-racing-problem}
    \begin{align}
        \min_{T, u_0, \ldots, u_{T-1} } & \quad \sum_{t = 0}^{T-1} 1 \\
        \text{s.t. } & \quad x_{t+1} = f(x_t, u_t), \quad\forall t = [0,\dots, T-1] \label{eq:lmpc_dynamics}\\
        & \quad x_t \in \mathcal{X},\ u_t\in \mathcal{U},\quad\forall t = [0,\dots, T] \label{eq:lmpc_constraints}\\
        & \quad x_T \in \mathcal{X}_F,~ x_0 = x_S,
    \end{align}
\end{subequations}
where \eqref{eq:lmpc_dynamics} represents the system dynamics model, and \eqref{eq:lmpc_constraints} are the constraints of system states and inputs.
For each iteration, the system starts at the starting point $x_S$ and is driven to reach the terminal set $\mathcal{X}_F$.
For instance, in an autonomous racing competition, $x_S$ represents the starting line of the track;
$\mathcal{X}_F = \{x \in \mathbb{R}^6 : [0~0~0~0~1~0] x = s_c \geq L \}$ represents the states beyond the finish line when the length of the track is $L$.

For each $j$-th iteration that successfully drives the system to the terminal set $\mathcal{X}_F$, the algorithm stores the historical states and inputs in a historical data set $\mathcal{H}$.
Meanwhile, after the completion of $j$-th iteration, cost-to-go values, representing the time to finish the corresponding iteration from the stored state to the terminal set, are computed and stored for each stored historical state.
Here, we discuss the LMPC algorithm for car racing competitions as described in \cite{rosolia2019learning}.
For this scenario, the cost-to-go represents the time to finish the corresponding lap from the state to the finish line and is computed after the ego racing car finishes the corresponding lap.

To formulate the local optimization problem, a subset of the stored historical states is used to construct a local convex set, which is used as the terminal constraint of the local optimization problem.
Since this convex set includes the states that drive the ego vehicle to the finish line in the previous laps, this terminal constraint formulation guarantees the feasibility of the system. 

For the car racing competition, this convex set is defined as the convex hull of the $K$-nearest neighbors to the current state $x_t$.
The following criterion is used to select these points:
\begin{subequations}\label{eq:nearest_points}
\begin{align}
    \argmin\limits_{{z}_g} 
    \sum\limits_{g=1}\limits^{K}||{z}_g - &{x}_t||^{2}_{D_z} \\
    \text{s.t.}\quad z_{g} \neq z_{m}, &~\forall{g\neq m}\\
     z_{g} \in \mathcal{H},&~g= 1,...,K 
\end{align}
\end{subequations}
where $\mathcal{H}$ is the historic data set, ${D_z}$ is a diagonal matrix that contains weighting coefficients for state variables.
Consequently, we select $K$-nearest points around current state $x_t$ at each timestep considering weighting $D_z$ for each dimension.
To enhance computational efficiency, only the historical states from iteration $l$ to $j$ are utilized for the selection of $K$-nearest points.
Then, the following matrix is used to store these $K$ points:
\begin{equation}
    D^j_l(x) = [x_{t^{l,*}_1}^l, \ldots, x_{t^{j,*}_K}^j]
\end{equation}
Then, the local convex safe set around $x_t^j$ can be defined as:
\begin{equation}
\begin{aligned}
\mathcal{CL}^j_l(x) = \{\bar{x} \in \mathbb{R}^6 : \exists \lambda \in &\mathbb{R}^{K}, \, \\
&\lambda \geq 0, \, \mathbf{1}^\top \lambda = 1, \, D_l^j(x)\lambda = \bar{x} \}
\label{eq:convex_set}
\end{aligned}
\end{equation}

Meanwhile, the following local convex Q-function is defined as the convex combination of the cost-to-go associated with the $K$-nearest neighbors we select:
\begin{equation}
\begin{aligned}
Q_j(\bar{x}, x) &= \min_{\boldsymbol{\lambda}} \, {\bf{J}}_j(x) \boldsymbol{\lambda} \\
\text{s.t.} \quad &~ \boldsymbol{\lambda} \geq 0, \mathbf{1}^\top \boldsymbol{\lambda} = 1, D^j_l(x) \boldsymbol{\lambda} = \bar{x}
\end{aligned}
\end{equation}
where $\boldsymbol{\lambda} \in \mathbb{R}^{K}$, $\mathbf{1}$ is a vector of ones and the row vector
\begin{equation*}
\begin{aligned}
    {\bf{J}}^j_l(x) = [J_{t^{l,*}_1 \rightarrow T^l}^l &(x_{t^{l,*}_1}^l), \ldots, J_{t^{j,*}_K\rightarrow T^j}^j(x_{t^{j,*}_K}^j)],
\label{vec:Q-function}
\end{aligned}
\end{equation*}
represents the cost-to-go associated with the $K$ historical states from the $l$-th to the $j$-th iteration.
The cost-to-go $J_{t \rightarrow T^j}^j (x_{t}^j) = T^j - t$ denotes the remaining time required for the vehicle to travel from state $x_t^j$ to the finish line along the $j$-th trajectory.


Then, for time step $t$ of $j$-th iteration, the LMPC algorithm aims to solve the following finite-time optimal control problem by using the local convex safe set and Q-function:
\begin{subequations}\label{eq:FTOCP}
\begin{align}
    J_{t\rightarrow t+N}^{j}(&x_t^j, z_t^j) = \notag \\
    \min_{{\bf{U}}_t^j, \boldsymbol{\lambda}_t^j}  \quad &\bigg[  \sum_{k=t}^{t+N-1}  h(x_{k|t}^j) +{\bf{J}}^{j-1}_l(z_t^j) \boldsymbol{\lambda}_t^j\bigg] \label{eq:FTOCP_Const}\\
	\text{s.t.}\quad &x_{t|t}^j=x_t^j, \label{eq:FTOCP_IC}\\
	&\boldsymbol{\lambda}_t^j \geq 0, \textbf{1}^\top{\boldsymbol{\lambda}}_t^j = 1,  D^{j-1}_l(z_t^j) \boldsymbol{\lambda}_t^j = x_{t+N|t}^j, \label{eq:FTOCP_Term}\\
    \quad &x_{t+1} = f(x_t, u_t), \label{eq:FTOCP_Dyn} \\
	&x_{k|t}^j \in \mathcal{X}, u_{k|t}^j \in \mathcal{U}, 
	\forall k = t, \cdots, t+N-1, \label{eq:FTOCP_Cons}
\end{align}
\end{subequations}
where $N$ denotes the prediction horizon for the controller, $x_t^j$ represents the current state, $\boldsymbol{\lambda}_t^j \in \mathbb{R}^{K}$ is used to describe the local convex Q-function defined in \eqref{vec:Q-function}.
In the cost function \eqref{eq:FTOCP_Const}, the stage cost $h(.)$ is defined as
\begin{equation}
\label{eq:stage cost}
    h(x) = \begin{cases} 1 & \mbox{If } x \notin \mathcal{X}_F\\
    0 & \mbox{Else }\\
    \end{cases}.
\end{equation}

$z_t^j$ is a candidate terminal state.
It is selected as the convex combination of the columns of the matrix $S_l^j(x)$, which is the evolution of the states stored in the columns of the matrix $D^j_l(x)$: $ S_l^j(x) = [x_{t^{l,*}_1+1}^l, \ldots,  x_{t^{j,*}_K+1}^j]$.
The constraint \eqref{eq:FTOCP_Term} guarantees that the terminal point $x_{t+N|t}^j$ lies within the convex set $\mathcal{CL}^j_l(z_t^j)$.
Equation \eqref{eq:FTOCP_Dyn} describes the vehicle dynamics, while equation \eqref{eq:FTOCP_Const} represents the state and input constraints.

Since the cost function is based on the cost-to-go associated with stored historical states, this local optimization problem guarantees that the vehicle can reach the finish line with time that is no greater than the time from the same position during previous laps.
In this way, the algorithm improves the racing car's lap time performance continuously and reaches its time-optimal performance after several laps.


The objective of the optimization problem at timestep $t$ is to minimize the cost associated with reaching the terminal state, while ensuring reachability.
This is achieved by finding the optimal vector $\boldsymbol{\lambda}_t^j$ within the convex set constructed based on the potential terminal state $z_{t}^j$, which determines the terminal state 
$x_{t+N|t}^j$ that the system seeks to reach from $x_t^j$.
Since the cost function is based on the cost-to-go associated with stored historical states, this local optimization problem guarantees that the vehicle can reach the finish line with time that is no greater than the time from the same position during previous laps.
In this way, the algorithm improves the racing car's lap time performance continuously and reaches its time-optimal performance after several laps.

Based on the LMPC algorithm, three autonomous racing strategies are developed to drive the ego racing car to compete with moving racing cars on the track.

\subsection{Learning-based MPC with Local re-Planning}\label{sec:appendix-re-planning}

To compete with multiple competitors in a car racing scenario, authors in \cite{he2022autonomous} propose an autonomous racing strategy that switches between two modes.
When there are no surrounding vehicles, the LMPC \eqref{eq:FTOCP} trajectory planner is used to improve the ego vehicle's lap time performance.
When the ego vehicle is overtaking leading vehicles, an optimization-based planner generates feasible trajectories and a low-level MPC controller is used to track them.
In this section, we primarily discuss the racing strategy when competing against other vehicles.

When leading vehicles are sufficiently close, an optimization-based trajectory planner is activated to optimize several homotopic trajectories in parallel.
Following optimization problem is used to compute these trajectories:
\begin{subequations}
\label{eq:traj_optimization}
\begin{align}
& \argmin_{x_{t:t{+}N_p|t}, u_{t:t{+}N_p{-}1|t}} p(x_{t{+}N|t}){+}\sum\limits_{k=0}^{N_p-1}q(x_{t{+}k|t})  
 \notag \\
& \quad {+}\sum\limits_{k=1}^{N_p-1}r(x_{t{+}k|t}, u_{t{+}k|t}, {x_{t{+}k{-}1|t}, u_{t{+}k{-}1|t}}) \label{eq:cost_planner} \\ 
\text{s.t.} \ & x_{t+k+1|t} = A x_{t+k|t} + B u_{t+k|t}, k= 0,...,N_p{-}1, \label{eq:dynamics_update}  \\ 
& x_{t+k+1|t} \in \mathcal{X}, u_{t+k|t} \in \mathcal{U}, k= 0,...,N_p{-}1, \label{eq:state_input_constraint} \\
& x_{t|t} = x_{t}, ~~~~~ \label{eq:initial_state} \\ 
& g(x_{t+k+1|t}) \geq d + \epsilon, k= 0,...,N_p{-}1, \label{eq:safety_constraint}
\end{align}
\end{subequations}
where \eqref{eq:dynamics_update}, \eqref{eq:state_input_constraint}, \eqref{eq:initial_state} are constraints for vehicle dynamics, state/input bounds and initial condition.
When there are $n$ surrounding vehicles, $n+1$ sets of Bézier curves are used as reference paths. 
The cost function \eqref{eq:cost_planner} generates $n+1$ trajectories by minimizing deviations from these reference curves while optimizing timing performance.
The feasible trajectories generated by this optimization are then evaluated, and the autonomous racing strategy selects the optimal solution from the multiple candidate trajectories, which is subsequently tracked by a low-level MPC controller.

\subsection{Learning-based MPC with Slacked Target State}\label{sec:appendix-slack-target}

In this section, to enhance feasibility, the authors in \cite{rosolia2021minimum} modify the LMPC formulation.
Specifically, they utilize a series of points from $K$-nearest neighbors $D^{j-1}_l(z_t^j)$ instead of the local convex hull in equation \eqref{eq:FTOCP_Term}.
For each candidate point $z_g \in D^{j-1}_l(z_t^j)$, they solve an optimization problem in ascending order based on the cost-to-go.
The optimal control problem for each $z_g$ is defined as follows:

\begin{subequations}\label{eq:lmpc-slack-target}
\begin{align}
    J_{t\rightarrow t+N}^{j}(&x_t^j, z_g) = \notag \\
    \min_{{\bf{U}}_t^j, \boldsymbol{\lambda}_t^j}  \quad &\bigg[  \sum_{k=t}^{t+N-1}  h(x_{k|t}^j) +\xi_1^\top \mathbf{Q}_{\text{slack},1} \xi_1\bigg] \label{eq:lmpc-slack-target-Const}\\
	\text{s.t.}\quad &x_{t|t}^j=x_t^j, 
    \label{eq:lmpc-slack-target-IC}\\
	&x_{t+N|t}^j + \xi_1 = z_g, z_g\in D^{j-1}_l(x_t^j)\label{eq:lmpc-slack-target-Term}\\
    \quad &x_{t+1} = f(x_t, u_t) \quad, 
    \label{eq:lmpc-slack-target-Dyn} \\
	&x_{k|t}^j \in \mathcal{X}, u_{k|t}^j \in \mathcal{U}, \label{eq:lmpc-slack-target-Cons}\\ 
	&\forall k = t, \cdots, t+N-1, \notag
\end{align}
\end{subequations}

In the above equations, the slack vector $\xi_1 \in \mathbb{R}^{6}$ is introduced to provide flexibility in the system's terminal state constraints.
$\xi_1$ allows for small deviations between $z_g$ and $x_{t+N|t}^j$, thereby ensuring the problem remains feasible even if the ideal solution cannot be perfectly achieved.
Additionally, $\xi_1^\top \mathbf{Q}_{\text{slack},1} \xi_1$ is added in the equation \eqref{eq:lmpc-slack-target-Const} to balance feasibility and optimality.
$\mathbf{Q}_{\text{slack},1} \in \mathbb{R}^{6 \times 6}$ is the weight matrix that allows for a certain degree of constraint violation.
By adding obstacle avoidance constraint to the optimization problem, this formulation is suitable to compete with other racing cars on the race track.

\subsection{Learning-based MPC with Slack on Convex Hull}\label{sec:appendix-slack-hull}

In this section, we introduce another method for Relaxed LMPC in \cite{rosolia2021minimum}.
The authors build upon the LMPC problem \eqref{eq:FTOCP} by relaxing the convex hull constructed in equation \eqref{eq:FTOCP_Term} with the addition of slack variable $\xi_2$. 
Thus, the optimal control problem is updated as follows:

\begin{subequations}\label{eq:lmpc-slack-ch}
\begin{align}
    J_{t\rightarrow t+N}^{j}(&x_t^j, z_t^j) = \notag \\
    \min_{{\bf{U}}_t^j, \boldsymbol{\lambda}_t^j}  \quad &\bigg[  \sum_{k=t}^{t+N-1}  h(x_{k|t}^j) + {\bf{J}}^{j-1}_l(z_t^j) \boldsymbol{\lambda}_t^j + \mathbf{Q}_{\text{slack},2} \cdot \xi_2^2\bigg] \label{eq:lmpc-slack-ch-Const}\\
	\text{s.t.}\quad &x_{t|t}^j=x_t^j, 
    \label{eq:lmpc-slack-ch-IC}\\
    &\textbf{1}^\top{\boldsymbol{\lambda}}_t^j = 1 + \xi_2,  D^{j-1}_l(z_t^j) \boldsymbol{\lambda}_t^j = x_{t+N|t}^j \notag \\
    &\boldsymbol{\lambda}_t^j \geq 0, \xi_2 \geq 0,\label{eq:lmpc-slack-ch-Term}\\
    \quad &x_{t+1} = f(x_t, u_t) \quad, 
    \label{eq:lmpc-slack-ch-Dyn} \\
	&x_{k|t}^j \in \mathcal{X}, u_{k|t}^j \in \mathcal{U}, \label{eq:lmpc-slack-ch-Cons}\\ 
	&\forall k = t, \cdots, t+N-1, \notag
\end{align}
\end{subequations}
where $\xi_2$ is introduced as a scalar relaxation that permits minor deviations, allowing the optimization problem to remain feasible while minimizing the impact through the penalization term $\mathbf{Q}_{\text{slack},2} \cdot \xi_2^2$.
$\mathbf{Q}_{\text{slack},2}$ is the weighting factor that allows for a degree of relaxation of the convex hull.
\begin{remark}
Notice that in Appendices~\ref{sec:appendix-re-planning}~\ref{sec:appendix-slack-target} and~\ref{sec:appendix-slack-hull}, the obstacle avoidance constraints of the controller are implemented using discrete control barrier functions, expressed as follows:
\begin{equation}
b (x_{k+1|t}) \geq \gamma \omega_{k} b(x_{k|t}), \ k = t,...,t+N
\end{equation}
$\gamma \in [0, 1)$ and $\omega_{k}$ are tunable parameters, $N$ is the prediction horizon for the controller and  $b(.)$ indicates the safe boundary which is presented in \eqref{eq:safe_boundary}.
In \eqref{eq:safe_boundary}, the safe boundary relies on the predicted trajectories of surrounding vehicles. These trajectories are deterministic, as we pre-generate the trajectories of all surrounding vehicles within a specified simulation timeframe.
Consequently, our control strategy and these baseline strategies can obtain the predicted trajectories, which are of horizon $N$, based on the current simulation time of the ego vehicle.
\end{remark}

\begin{remark}
    The two slack formulations \eqref{eq:lmpc-slack-target-Const} \eqref{eq:lmpc-slack-ch-Term} serve distinct geometric purposes: While in Appendix~\ref{sec:appendix-slack-target} the equation \eqref{eq:lmpc-slack-target-Const} relaxes the terminal state $x_{t+N|t}^j$ to exactly match an individual historical point $z_g \in D^{j-1}_l(x_t^j)$ via $\xi_1$ (bypassing convex hull constraints entirely), the formulation \eqref{eq:lmpc-slack-ch-Term} in Appendix~\ref{sec:appendix-slack-hull} instead introduces relaxation on the convex weights $\boldsymbol{\lambda}_t^j$ via $\xi_2$, allowing the terminal state to deviate slightly from the hull's interior. 
    The former prioritizes exact convergence to specific historical states, whereas the latter maintains approximate adherence to the historical distribution's shape.
\end{remark}

\vfill

\end{document}